\documentclass[sigconf]{acmart}
\usepackage{bm}
\usepackage[ruled,vlined]{algorithm2e}
\usepackage[noabbrev, capitalize]{cleveref}
\usepackage{pifont, xspace}
\usepackage{enumitem}
\AtBeginDocument{%
  }

\copyrightyear{2026}
\acmYear{2026}
\setcopyright{cc}
\setcctype{by}
\acmConference[SIGSPATIAL '26]{The 34th ACM International Conference on Advances in Geographic Information Systems}{November 03--06, 2026}{Riverside, CA, USA}
\acmBooktitle{The 34th ACM International Conference on Advances in Geographic Information Systems (SIGSPATIAL '26), November 03--06, 2026, Riverside, CA, USA}
\acmDOI{10.1145/3841645.3843313}
\acmISBN{979-8-4007-2950-8/2026/11}

\begin{document}

\title[\genesis{}: Hierarchical Satellite Image Generation]{Genesis: A \underline{Gen}erative \underline{E}ngine for Hierarchical \underline{S}atellite \underline{I}mage \underline{S}ynthesis}

\author{Subash Khanal}
\affiliation{%
  \institution{Washington University in St. Louis}
  \city{Missouri, USA}
  \country{}}

\author{Yangzhi Cui}
\affiliation{%
  \institution{Washington University in St. Louis}
   \city{Missouri, USA}
  \country{}}

\author{Daniel Cher}
\affiliation{%
  \institution{Washington University in St. Louis}
   \city{Missouri, USA}
  \country{}}

\author{Eric Xing}
\affiliation{%
  \institution{Washington University in St. Louis}
  \city{Missouri, USA}
  \country{}}

\author{Brian Wei}
\affiliation{%
  \institution{Washington University in St. Louis}
  \city{Missouri, USA}
  \country{}}

\author{Srikumar Sastry}
\affiliation{%
  \institution{Washington University in St. Louis}
  \city{Missouri, USA}
  \country{}}

\author{Nathan Jacobs}
\authornote{Corresponding author: \texttt{jacobsn@wustl.edu}}
\affiliation{%
  \institution{Washington University in St. Louis}
   \city{Missouri, USA}
  \country{}}

\renewcommand{\shortauthors}{Khanal et al.}

\newcommand{\cmark}{\ding{51}}
\newcommand{\xmark}{\ding{55}}
\newcommand{\densedataset}{\textsc{dense500}\xspace}
\newcommand{\genesis}{\textsc{Genesis}\xspace}

\begin{abstract}
Earth observation is fundamentally multi-scale; geospatial tasks span varied resolutions, and
satellite imagery is organized into cascading tile pyramids that nest fine detail within wide
coverage. Current generative models of satellite imagery, however, operate along a single
axis: they either zoom to enhance a single tile's resolution or pan to extend imagery at a
fixed scale. As a result, no existing method produces a complete pyramid that stays consistent
across both scale and space, where a high-zoom tile must agree with the coarse context it
refines and with the neighbors it meets. Motivated by this gap, we introduce a new task,
\emph{multi-scale tile completion}: given a sparse set of seed tiles at arbitrary zoom levels
and positions, synthesize a complete, uniform quadtree that is globally consistent across both
scale and space. We approach this task with \genesis, a generative engine that brings both axes together by
composing two specialized operators over the quadtree, a vertical super-resolution model and a
horizontal mask-based outpainting model, producing pyramids that are consistent across zoom
levels and seamless across neighboring tiles. Each operator achieves state-of-the-art results
on its subtask, and the engine propagates sparse seeds into seamless, multi-resolution maps
from any initial configuration. To evaluate the task and benchmark \genesis, we introduce
\densedataset, a fully observed multi-scale pyramid dataset spanning diverse geographic
regions, together with a suite of pyramid-level metrics. Code, models, and our dataset are available at \url{https://github.com/mvrl/genesis}.
\end{abstract}



\begin{teaserfigure}
\centering
\includegraphics[width=.618\linewidth]{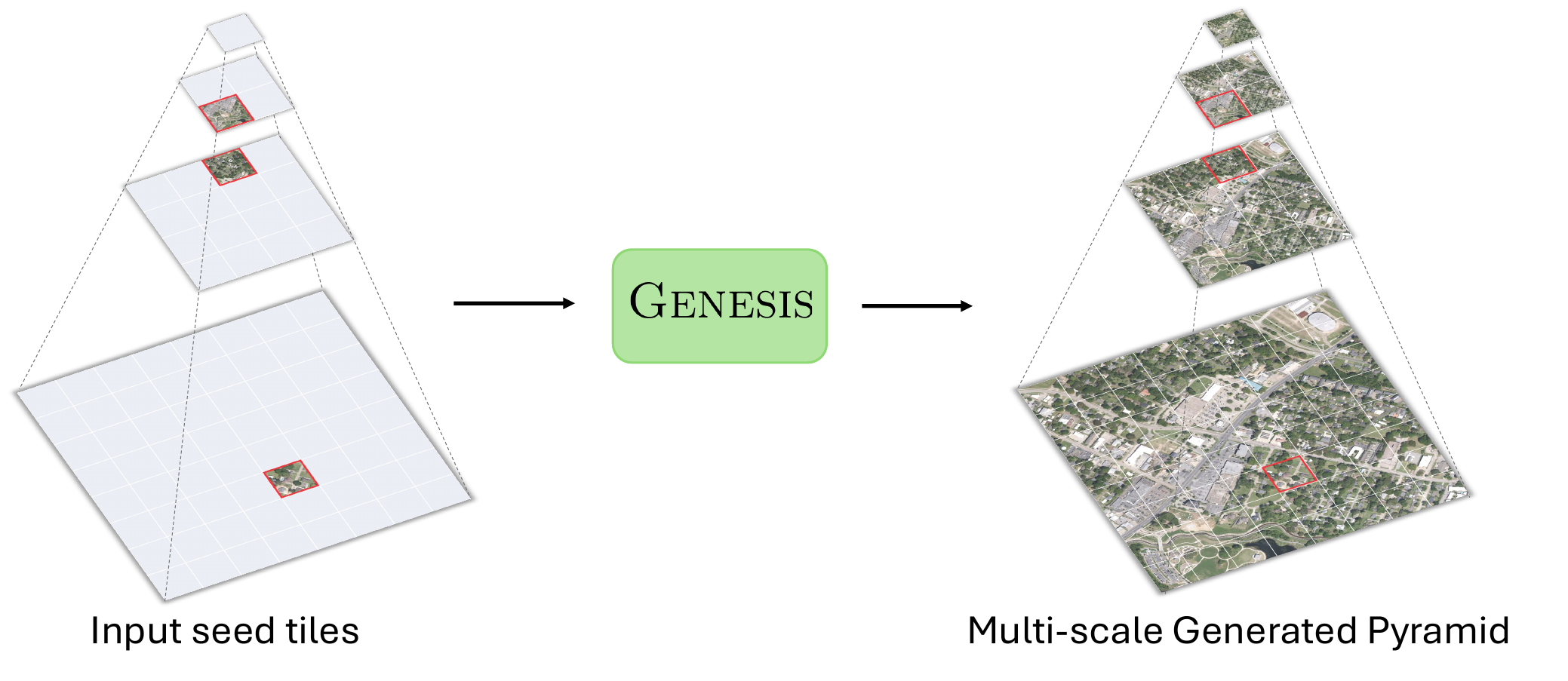}
\caption{Given sparse seed tiles (\textcolor{red}{red}) at arbitrary positions and zoom
levels, \textsc{Genesis} completes the entire pyramid (a uniform quadtree), such that the output
preserves the seeds, is seamless across neighboring tiles, and is consistent across zoom levels.}
\label{fig:task}
\end{teaserfigure}

\maketitle
\section{Introduction}
\label{sec:introduction}

Earth observation spans many scales. A task may concern the land cover of an entire continent or the footprint of a single building, and the resolution it demands shifts accordingly. To accommodate this range, satellite imagery is conventionally served as a tile pyramid: a hierarchy of zoom levels in which each tile subdivides into finer children at the level below. A generative model intended to synthesize the Earth's surface should therefore produce not an isolated image, but a coherent pyramid spanning many resolutions at once.

Existing generative models of satellite imagery address only a single axis of this structure. Outpainting and inpainting methods~\cite{rombach2022high,liu2025text2earth}, along with MultiDiffusion-style approaches~\cite{bar2023multidiffusion, lee2023syncdiffusion, jimenez2023mixture} that tile a denoiser to extend an image, operate \emph{horizontally}, growing content at a fixed zoom level. Super-resolution methods~\cite{yellapragada2025zoomldm,meng2024fastdiffsr,liang2021swinir} operate \emph{vertically}, refining a single tile across successive zoom levels. In each case, the other direction is held fixed. Yet generating a complete multi-resolution world requires moving vertically and horizontally at once: keeping a high-zoom tile consistent with the coarse context it refines, and keeping neighbors consistent where they meet. This joint constraint remains an open challenge.

We formalize this problem as \emph{multi-scale tile completion}. Given a sparse set of seed tiles placed at arbitrary zoom levels and positions, the objective is to reconstruct the complete tile pyramid they imply, populating every tile at every scale subject to two consistency constraints: vertical agreement between a parent and its children, and horizontal continuity between adjacent tiles. Such a capability converts sparse, opportunistic acquisitions into dense, navigable maps, with applications in populating virtual environments for games and simulators~\cite{panagiotou2020procedural}, exploring hypothetical urban layouts~\cite{wang2025generative}, and generating multi-resolution training data for remote sensing~\cite{he2020sat2graph,hetang2024segment,li2020object}.

To address this task, we introduce \textsc{\textbf{\genesis}}, a \underline{\textbf{Gen}}erative \underline{\textbf{E}}ngine for Hierarchical \underline{\textbf{S}}atellite \underline{\textbf{I}}mage \underline{\textbf{S}}ynthesis. \genesis decomposes the task into two complementary generative operators defined over the quadtree. A vertical operator performs super-resolution, mapping an $N{\times}N$ tile at zoom $z$ to the $2N{\times}2N$ mosaic of its four children at $z{+}1$. A horizontal operator performs mask-based outpainting, completing a tile from an arbitrary subset of its known quadrants. Together with deterministic downsampling, these operators provide all the tools needed to complete the pyramid: any seed configuration can be propagated to any other tile. Each model attains state-of-the-art performance on its respective subtask, and \genesis composes them to expand a sparse set of seeds into a seamless, multi-resolution map.

Multi-scale tile completion is a new task, and we introduce it together with a benchmark and evaluation protocol to support future research. We release \densedataset{}, a fully observed pyramid benchmark spanning multiple geographic regions and zoom levels, together with pyramid-level metrics to quantitatively assess generation quality. We hope this task, benchmark, and evaluation protocol will serve as a foundation for the community to explore multi-scale generative modeling of the physical world.

\paragraph{Our key contributions are:}
\begin{itemize}
    \item \textbf{A new task, \emph{multi-scale tile completion}:} generating a complete multi-resolution tile pyramid from a sparse set of seed tiles at arbitrary zoom levels, requiring both vertical consistency across zoom levels and horizontal consistency within a level.
      
    \item \textbf{The \genesis engine:} a unified framework that synthesizes a full pyramid
    from any seed configuration by composing two state-of-the-art satellite image generative models for super-resolution and outpainting.
    
    \item \textbf{A benchmark and evaluation protocol:} \densedataset, a fully observed pyramid benchmark spanning multiple geographic regions and zoom levels, together with pyramid-level metrics to quantitatively assess multi-scale generation quality.
\end{itemize}

\section{Related Work}
\label{sec:related_works}

Image generation has progressed rapidly, both in general and for satellite imagery specifically. Building large, coherent scenes for satellite imagery further draws on two spatial operators, super-resolution and outpainting, which refine content across scale and extend it across space, respectively. Reaching a wide-area extent at high resolution requires composing the latter across multi-tile regions.

\subsection{Generative Models of Satellite Imagery}
Modern image generation pipelines rest on diffusion models~\cite{ho2020ddpm, song2021score}, which have surpassed earlier GAN-based approaches~\cite{goodfellow2014gan}. Latent diffusion~\cite{rombach2022high} made high-resolution synthesis tractable by operating in a compressed VAE space, and subsequent work improved further by exploiting the scaling properties of transformer backbones~\cite{peebles2023dit, ma2024sit}, often joined with flow-based training~\cite{lipman2023flowmatching,liu2023rectifiedflow}. Interestingly, recent work~\cite{wang2026pixnerd, lei2025there, yu2026pixeldit, chen2025pixelflow} shows that stronger fidelity comes from stepping back from the VAE entirely and denoising the image tokens directly in pixel space, avoiding the information bottleneck and reconstruction artifacts that latent-space models introduce. \genesis adopts the \textsc{JiT} (Just image Transformer) \cite{li2026back} structure, which demonstrated this principle with a plain Vision Transformer trained under a flow-matching objective, achieving state-of-the-art generation without a latent tokenizer or large-scale pre-training. 

With advancements in general-domain image generation, satellite image generation has followed suit to enable stronger generation with more diverse control.  This line of work has explored the inductive biases of overhead imagery that differ from those of natural images in perspective, scale, spectral content, and the metadata available
for conditioning, and has accordingly developed conditioning principles suited to the domain. Beyond text-to-image generation~\cite{xu2023txt2img, liu2025text2earth, pan2025earthsynth}, these include conditioning on geospatial metadata such as geolocation~\cite{sastry2024geosynth}, acquisition time, and ground-sampling distance~\cite{khanna2024diffusionsat}. A more recent thread grounds generation on instance-level semantics, conditioning on spatially localized text through vector geometry or sparse point queries~\cite{cher2026vectorsynth, sastry2026geodit, wei2026terraditomega}.
These works share a focus on the conditioned generation of a single tile. \genesis instead targets the generation of entire scenes,
expanding from a handful of seeds to the many tiles that compose a city, consistent across both space and scale.

\subsection{Super-resolution}
Refining a tile across zoom levels corresponds to single-image super-resolution (SISR). The field has progressed from early CNN-based mappings \cite{dong2015srcnn} through perceptual and adversarial methods \cite{wang2018esrgan} to attention-based transformers. SwinIR \cite{liang2021swinir} showed that windowed self-attention captures the long-range dependencies needed for sharp reconstruction and remains a strong perceptual baseline. Diffusion-based methods have since pushed perceptual quality further, casting super-resolution as conditional or zero-shot restoration \cite{saharia2023sr3, wang2024stablesr, yue2023resshift}. In the remote-sensing domain, FastDiffSR \cite{meng2024fastdiffsr} accelerates diffusion SR with a lightweight backbone, and ZoomLDM \cite{yellapragada2025zoomldm}, the work closest to our setting, conditions a latent diffusion model on zoom level for scale-aware satellite synthesis. Yet all of these methods operate on a single tile in isolation. They add detail one step at a time and provide no mechanism to keep a refined tile consistent with the coarse parent it descends from, or with the neighboring tiles it must seamlessly join.

\subsection{Outpainting and Multi-tile Generation}
Extending content within a fixed zoom level corresponds to image inpainting and outpainting, which hallucinate plausible content for masked or out-of-frame regions conditioned on visible context. The task evolved from GAN-based context encoders \cite{pathak2016contextencoders} and large-mask convolutional methods \cite{suvorov2022lama} to diffusion-based approaches that offer markedly higher fidelity \cite{lugmayr2022repaint, saharia2022palette, rombach2022high}, and the same machinery has been adapted to complete aerial scenes under text conditioning \cite{liu2025text2earth}. These methods complete a single image. Covering a spatial \emph{region} larger than one tile instead requires piecing multiple tiles together, which tiled and multi-stage diffusion strategies \cite{bar2023multidiffusion, ho2022cascaded} achieve by fusing overlapping or stacked generations into seamless larger images. Yet this stitching happens at a single resolution. It maintains horizontal coherence between neighbors but has no notion of the coarser context a tile must refine or the finer detail it must contain, the cross-scale agreement that fuses a stack of tiles into a consistent pyramid.

\section{Data}
\label{sec:data}

We train our models on the Git-10M satellite imagery corpus~\cite{liu2025text2earth}. Git-10M is a sparsely sampled global tile grid that lacks the fully observed pyramids needed to score multi-scale completion. We therefore introduce \densedataset, a benchmark of $500$ fully observed depth-$4$ quadtrees released with this work. Both datasets consist of $256{\times}256$ tiles sampled in a quadtree structure where finer zooms double the ground resolution. 

\subsection{Git-10M Training Data}
\label{sec:data:git10m}

Git-10M~\citep{liu2025text2earth} provides $\sim$$10M$ satellite tiles at zoom levels
$10$ through $18$ with near-global coverage. Of these, $\sim$$7M$ carry Web-Mercator
coordinates and participate in the quadtree hierarchy; the remainder are not
georeferenced and serve only as auxiliary training images. Most georeferenced tiles lack
some of their relatives, so we reconstruct the partial hierarchy by linking available
parent-to-child pairs.

Evaluation quads are held out whole, so a parent and its four children
are always held out together, and any tile within an evaluation subtree is excluded from training, ensuring
no test content leaks into the training set. Our spatial test split equally samples urban
and non-urban quads, where urban is defined as the tile center lying within a World Urban
Areas~\cite{world_urban_areas} polygon, enabling evaluation across land-use types,
resolutions, and locations worldwide. Details of our dataset splits are provided in \cref{sec:appendix:splits}.

\subsection{The \densedataset{} Benchmark}
\label{sec:data:dense500}
 
\densedataset comprises $500$ fully observed depth-$4$ quadtrees, each $85$ tiles ($1{+}4{+}16{+}64$ at four consecutive zooms), organized as five zoom windows of $100$ geographically distinct sites (\cref{tab:dense500}). Because every tile of every site is observed, any seed configuration can be completed, and the result scored against ground truth at every level.

\begin{table}[t]
\centering
\small
\caption{The five zoom windows of \densedataset. Each window slides a four-level span up the pyramid, probing completion from roughly $76$ m/px down to $0.6$\,m/px.}
\label{tab:dense500}
\begin{tabular}{lccccc}
\toprule
Window & W1 & W2 & W3 & W4 & W5 \\
\midrule
Root zoom $z$       & $11$ & $12$ & $13$ & $14$ & $15$ \\
Leaf zoom $z{+}3$   & $14$ & $15$ & $16$ & $17$ & $18$ \\
Root footprint (km) & $18$ & $9$ & $4$ & $2$ & $1$ \\
Sites               & $100$ & $100$ & $100$ & $100$ & $100$ \\
\bottomrule
\end{tabular}
\end{table}

Sites are sampled per window with probability proportional to population density, restricted to urban cells, continent-stratified, and spaced at least $80$\, km apart. Candidates intersecting any Git-10M location are
discarded so the benchmark is disjoint from training. A final filter drops any site whose imagery source shifts across an adjacent zoom pair, most often the satellite-to-aerial transition that occurs at higher zoom levels, as measured by per-level color-distribution distance. Each window is capped at exactly $100$ sites. Further details on site selection and filtering are provided in \cref{sec:appendix:dense500}. \densedataset intentionally emphasizes populated regions, where fine-zoom imagery is densest, and demand is highest; the operator evaluations complement it by covering urban and non-urban tiles equally via the spatial split (\cref{sec:data:git10m}).

\providecommand{\bm}[1]{\boldsymbol{#1}}
\providecommand{\xclean}{\bm{x}}          %
\providecommand{\enoise}{\bm{\epsilon}}   %
\providecommand{\vvel}{\bm{v}}            %
\providecommand{\xnoisy}{\tilde{\bm{x}}}  %
\providecommand{\netfn}{\operatorname{net}}
\providecommand{\etal}{et al.\@}

\begin{figure*}[t!]
    \centering
    \includegraphics[width=0.8\linewidth]{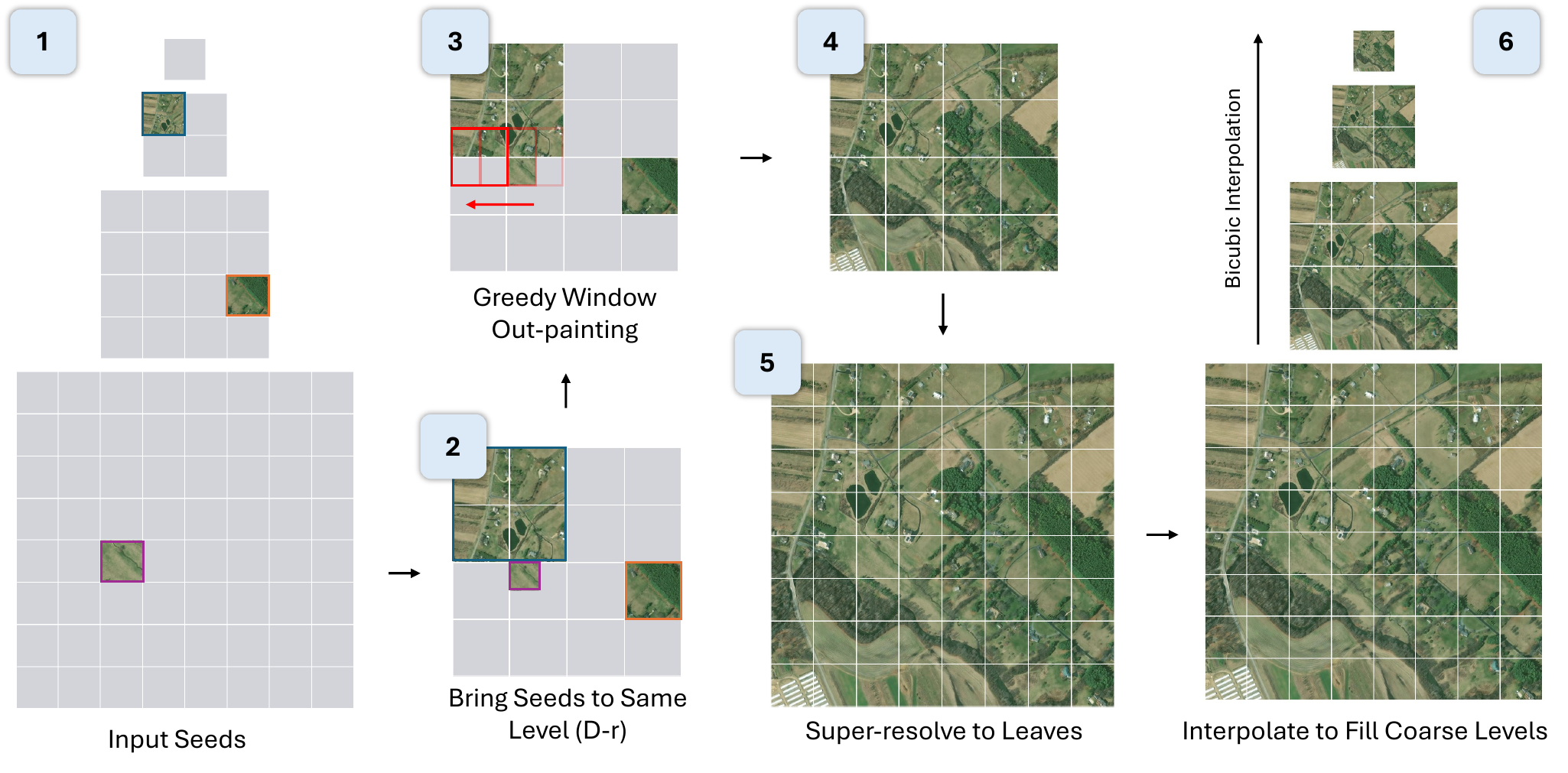}
    \caption{A high-level overview of the \genesis framework.}
    \label{fig:arch}
\end{figure*}

\section{Method}
\label{sec:method}

\genesis enables the generation of arbitrary geospatial extents given a set of sparse seed tiles. We now formalize this task and describe our method.

\subsection{Multi-scale Tile Completion}
\label{sec:method:task}

A tile pyramid is a uniform, complete quadtree (\cref{fig:task}). A tile $t=(z,x,y)$ at zoom level $z$
is an $N{\times}N$ image; it has a single parent at $z-1$ and four children at
$z+1$ --- its $2{\times}2$ quadrants, each doubling resolution. Given a set of
\emph{seed} tiles $\mathcal{S}$ at arbitrary zoom levels, the task is to fill
every remaining tile of a target subtree $\mathcal{T}$ so that the result is
consistent both \emph{vertically} (a parent equals the downsampled mosaic of its
children) and \emph{horizontally} (adjacent tiles join seamlessly).
We solve this with three operators over the quadtree:
\begin{itemize}
\item \textbf{Super-resolution (vertical, downward).} A generative model that maps an
$N{\times}N$ tile at $z$ to the stitched $2N{\times}2N$ mosaic of its four
children at $z+1$ (\cref{sec:method:sr}).
\item \textbf{Downsample (vertical, upward).} The deterministic $\div2$ inverse of super-resolution (SR), implemented by bicubic downsampling. A parent is an $N{\times}N$ tile at $z$, obtained as the downsampled mosaic of its four children at $z+1$.
\item \textbf{Outpainting (horizontal).} A generative model that completes an
$N{\times}N$ tile from an arbitrary subset of its known quadrants
(\cref{sec:method:op}).
\end{itemize}

\cref{sec:method:engine} and \cref{alg:genesis} specify how these operators are scheduled over the quadtree, and \cref{sec:exp:engine} gives the system-level instantiation.

\subsection{Pixel-space Generative Transformer}
\label{sec:method:jit} 
Both operators of \genesis are independently trained
\textsc{JiT} (Just image Transformer)~\cite{li2026back} models---each a
flow-matching transformer that operates directly in pixel space. In this section,
we therefore provide a brief review of \textsc{JiT} before describing how
\genesis adapts it to each operator. Throughout, we write $\tau$ for the
flow time, with $\tau{=}1$ corresponding to clean data and $\tau{=}0$ to noise.

\paragraph{Flow matching.}
Given a clean image $\xclean$ and Gaussian noise
$\enoise\sim\mathcal{N}(\mathbf{0},\mathbf{I})$, we define the linear interpolation
\begin{equation}
\xnoisy_\tau = \tau\,\xclean + (1-\tau)\,\enoise,
\qquad \tau\in[0,1],
\label{eq:interp}
\end{equation}
so that $\xnoisy_\tau$ recovers the clean image at $\tau{=}1$ and pure noise at $\tau{=}0$.
The associated target velocity is the time derivative of \cref{eq:interp}.
\begin{equation}
\vvel = \dot{\xnoisy}_\tau = \xclean - \enoise.
\label{eq:vtarget}
\end{equation}
During training, $\tau$ is drawn from a logit-normal distribution,
\begin{equation}
\mathrm{logit}(\tau)\sim\mathcal{N}(\mu,\,\sigma^2).
\label{eq:logitnormal}
\end{equation}

\paragraph{Prediction space and loss space.}
As \citet{li2026back} observe, the space in which the network makes its prediction
and the space in which the loss is applied can be decoupled. \genesis
exploits this with the $\xclean$-prediction / $\vvel$-loss combination: the network
$\mathrm{net}_\theta$ directly predicts the clean image,
\begin{equation}
\xclean_\theta = \mathrm{net}_\theta(\xnoisy_\tau,\tau),
\label{eq:xpred}
\end{equation}
which is mapped to a velocity prediction through the same interpolation,
\begin{equation}
\vvel_\theta = \frac{\xclean_\theta - \xnoisy_\tau}{1-\tau}.
\label{eq:x2v}
\end{equation}
The choice of $\xclean$ as the \emph{prediction} space is what makes training directly
in pixels viable: \citet{li2026back} argue that clean data lie on a low-dimensional
manifold, while a noised quantity is distributed across the full high-dimensional
space, and show empirically that $\enoise$- and $\vvel$-prediction degrade as the
dimension grows whereas $\xclean$-prediction remains effective. Following
\cite{li2026back}, this prediction is supervised in \emph{velocity} space, minimizing
\begin{equation}
\mathcal{L} = \mathbb{E}_{\tau,\xclean,\enoise}
\left\lVert \vvel_\theta - \vvel \right\rVert^2,
\qquad \vvel=\xclean-\enoise,
\label{eq:vloss}
\end{equation}
which retains a well-behaved regression objective while the network output stays on the
image manifold.

\paragraph{Sampling.}
Generation integrates the learned velocity field as an ordinary differential equation,
\begin{equation}
\frac{\mathrm{d}\xnoisy_\tau}{\mathrm{d}\tau} = \vvel_\theta(\xnoisy_\tau,\tau),
\label{eq:ode}
\end{equation}
starting from $\xnoisy_0\sim\mathcal{N}(\mathbf{0},\mathbf{I})$ at $\tau{=}0$ and ending at
the generated image at $\tau{=}1$. We use a first-order Euler discretization with $K$ uniform
steps of size $\Delta\tau=1/K$:
\begin{equation}
\xnoisy_{\tau+\Delta\tau} = \xnoisy_\tau + \Delta\tau\,\vvel_\theta(\xnoisy_\tau,\tau).
\label{eq:euler}
\end{equation}

\subsection{Vertical Operator: \texorpdfstring{$\times2$}{x2} Super-resolution}
\label{sec:method:sr}
The super-resolution (SR) operator generates the $2N{\times}2N$ child mosaic of a
tile, conditioned on its parent. It follows the pixel-space transformer of
\cref{sec:method:jit}, with added parent conditioning.

\paragraph{Conditioning.}
The parent conditions generation through two token streams: pixel tokens from a patch-embedding of the parent, and semantic tokens from a frozen, satellite-pretrained \textsc{DINOv3} encoder. Both are injected through cross-attention at every block. The target zoom level $z$ is embedded and added to the AdaLN conditioning vector, $c = \mathrm{emb}_\tau(\tau) + \mathrm{emb}_z(z)$. Conditioning is jointly dropped during training to enable classifier-free guidance. Hyperparameters are detailed in \cref{sec:exp:operators}.

\paragraph{Training objective.}
The clean target $\xclean$ is the $2N{\times}2N$ mosaic stitched from the observed
children. Since seeds may supply only a subset of children, we restrict the velocity
loss (\cref{eq:vloss}) to the observed pixels via a binary mask $m$. We further add a
perceptual \textsc{LPIPS} loss on the directly predicted mosaic $\xclean_\theta$
(\cref{eq:xpred}), comparing each predicted child quadrant against its ground truth:
\begin{equation}
\mathcal{L}_{\mathrm{SR}}
= \underbrace{\mathbb{E}\!\left[\frac{\sum_p m_p\lVert\vvel_\theta-\vvel\rVert^2_p}{\sum_p m_p}\right]}_{\text{masked velocity}}
+\ \lambda_{\mathrm{perc}}\,
\underbrace{\frac{\mathbb{E}\!\left[\tau\sum_q m_q\,\mathrm{LPIPS}\!\left(\xclean_\theta^{(q)},\xclean^{(q)}\right)\right]}
{\mathbb{E}\!\left[\tau\sum_q m_q\right]}}_{\text{perceptual (LPIPS)}},
\label{eq:srloss}
\end{equation}
where $q$ indexes the four child quadrants and $m_q$ flags their presence; the LPIPS term is weighted by the flow time $\tau$, 
emphasizing near-clean samples (large $\tau$) where $\xclean_\theta$ is reliable. We optimize this objective under a \emph{zoom-level
curriculum} that progresses from coarse- to fine-parent pairs (\cref{sec:exp:backbone}), and ablate the use of \textsc{DINOv3} semantic features 
and \textsc{LPIPS} loss in \cref{tab:sr-ablation}.

\subsection{Horizontal Operator: Mask-based Outpainting}
\label{sec:method:op}
The outpainting (OP) operator completes a tile from an arbitrary subset of its known quadrants. It follows the pixel-space transformer of \cref{sec:method:jit}, with added conditioning on the known content. The region to synthesize is given by a binary mask $h\in\{0,1\}^{N\times N}$ ($h{=}1$ on hole pixels, $h{=}0$ on known pixels).

\paragraph{Conditioning.}
Noise is injected only inside the hole: the known region of $\xnoisy_\tau$ stays clean and \cref{eq:interp} is applied to the hole alone. Following standard inpainting practice, we form a seven-channel context by concatenating this partially noised tile $\xnoisy_\tau$, the mask $h$, and the masked clean tile $(1-h)\odot\xclean$. The context is patch-embedded into tokens and injected through cross-attention at every block, with attention restricted to the \emph{known} patches so generation is driven only by observed content. The target zoom level $z$ is embedded and added to the AdaLN conditioning vector as in SR (\cref{sec:method:sr}). Instantiation details are given in \cref{sec:exp:operators}.

\paragraph{Training objective.}
OP uses the masked velocity loss (the velocity term of \cref{eq:srloss}); here $h$ marks the hole, so the objective is evaluated only on the pixels to be synthesized,
\begin{equation}
\mathcal{L}_{\mathrm{OP}}
= \mathbb{E}\!\left[\frac{\sum_p h_p\lVert\vvel_\theta-\vvel\rVert^2_p}{\sum_p h_p}\right].
\label{eq:oploss}
\end{equation}
We train under a \emph{quadrant curriculum} that grows the hole monotonically. We start with three known quadrants, then two, then one, then a mixture (detailed in \cref{sec:exp:backbone}).

\paragraph{Inference.}
The known region is held fixed throughout sampling: after each Euler step
(\cref{eq:euler}) we reset it to the clean pixels,
\begin{equation}
\xnoisy_\tau \leftarrow h\odot\xnoisy_\tau + (1-h)\odot\xclean,
\label{eq:opfreeze}
\end{equation}
so that only the hole is generated while the observed quadrants are preserved exactly.

\subsection{The \genesis Engine}
\label{sec:method:engine}

\genesis composes the three operators to fill a target subtree of depth $D$ from any
seed set (\cref{alg:genesis}). It concentrates the generative work at a single
\emph{working level} $z_{D-r}$, chosen $r$ levels above the leaves: this level tiles
the whole subtree footprint at one resolution, so once it is complete, $r$ successive
SR passes refine it up to the leaves, and downsampling operations fill every
coarser level.

The offset $r$ balances the two generative operators. Outpainting at the working
level performs the lateral hallucination that extends the scene beyond the seeds. SR
is comparatively cheap, since each pass upsamples an $N{\times}N$ tile to its
$2N{\times}2N$ children at once, and mainly adds high-frequency detail. A coarser level
(larger $r$) shrinks the grid the outpainter must fill and leans more on SR
refinement; a finer level (smaller $r$, e.g.\ penultimate at $r{=}1$) shifts more
hallucination onto OP at high resolution.

\begin{figure*}[t!]
  \includegraphics[width=0.8\textwidth]{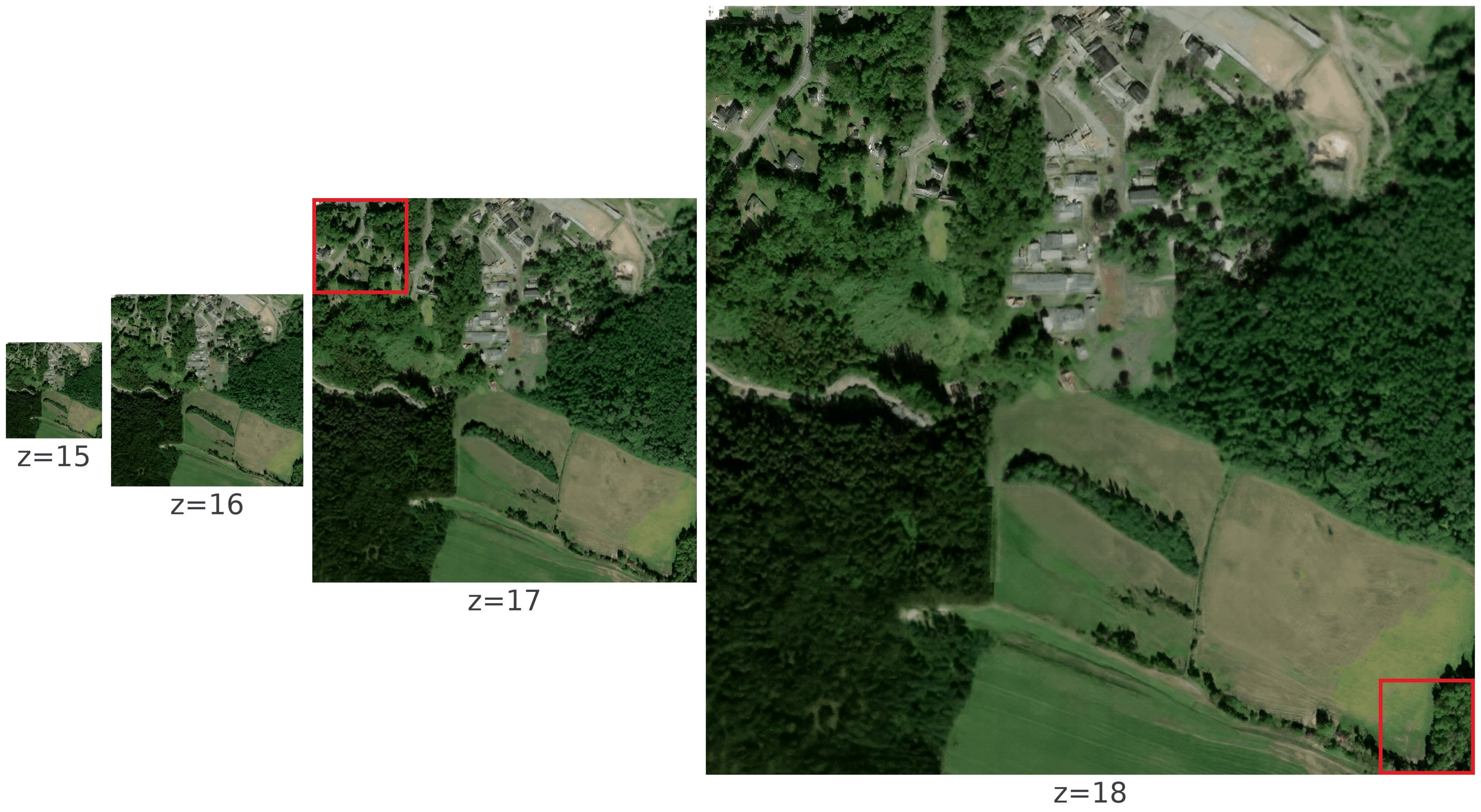}
\caption{From a set of seed tiles (\textcolor{red}{red}), \genesis synthesizes a complete
multi-scale pyramid, shown here as a progressive zoom of one site from $z{=}15$ to
$z{=}18$.}
  \label{fig:pyramid_example}
\end{figure*}

\begin{algorithm}[t]
\caption{\genesis pyramid completion}
\label{alg:genesis}
\KwIn{seeds $\mathcal{S}$; subtree $\mathcal{T}$ with leaf level $z_D$; offset $r$;
operators $\mathrm{SR}_\theta,\mathrm{OP}_\phi,\downarrow$}
\tcp{1. bring each seed to the working level $z_{D-r}$}
\ForEach{$s\in\mathcal{S}$}{
  \lIf{$z(s)<z_{D-r}$}{$\mathrm{SR}_\theta$ to $z_{D-r}$}
  \lElseIf{$z(s)=z_{D-r}$}{as-is}
  \lElse{$\downarrow$ to quadrant}
}
\tcp{2. maximum coverage outpainting on the quadrant grid}
\lWhile{a tile at $z_{D-r}$ has known and unknown quadrants}{$\mathrm{OP}_\phi$ completes the most-known one}
\tcp{3. SR up to the leaves, anchoring leaf seeds}
\lFor{$z = z_{D-r}$ \KwTo $z_D-1$}{level $z{+}1 \leftarrow \mathrm{SR}_\theta(\text{level }z)$}
\tcp{4. repair leaf seams with OP, worst first}
\lForEach{high-discontinuity seam}{$\mathrm{OP}_\phi$ regenerates and blends a band}
\tcp{5. reconcile by downsampling}
\lFor{$z = z_D-1$ \KwTo root}{non-seed tile $\leftarrow$ children$\,\downarrow$}
\Return{$\{I(t):t\in\mathcal{T}\}$}\; where $I(t)$
    denotes the generated image of tile $t$.
\end{algorithm}

Seeds are first brought to the working level (step~1): a coarser seed is
super-resolved, a finer seed is downsampled into its quadrant, and a seed already
at the level is used as-is. The rest of the level is completed by maximum coverage outpainting (step~2; detailed below). SR then refines the completed level up to the
leaves with any leaf seeds anchored in place (step~3). Any residual seams between 
adjacent leaf tiles are then repaired with OP, which regenerates a band across 
each high-discontinuity boundary and blends it back into the mosaic (step~4). 
Finally, a deterministic downsampling sweep fills every coarser level so that coarser
tiles stay consistent with the leaves (step~5); working-level tiles whose leaves were
untouched by seam repair are kept as-is. The result is a complete
pyramid that honors every seed and is consistent both vertically and horizontally.

\paragraph{Maximum coverage outpainting.}
Wherever \genesis applies OP, it greedily uses the maximum available tile coverage as contextual information. We treat the target mosaic as a grid of half-tiles (quadrants) and slide an
$N{\times}N$ outpainting window across it at half-tile offsets, so each window covers a
$2{\times}2$ block of quadrants and can straddle tile boundaries. Within a window, the
known quadrants condition OP, which completes the unknown one(s); we fill the windows
with the most known quadrants first, so every completion is anchored on as many sides
as possible. Because the windows overlap, each newly completed quadrant becomes
context for its neighbors, and content propagates outward from the seeds. Step~2
applies this regime to complete the working level, and the residual seam repair reuses the same windowed OP over leftover tile boundaries.

\providecommand{\bm}[1]{\boldsymbol{#1}}
\providecommand{\xclean}{\bm{x}}          %
\providecommand{\enoise}{\bm{\epsilon}}   %
\providecommand{\vvel}{\bm{v}}            %
\providecommand{\xnoisy}{\tilde{\bm{x}}}  %
\providecommand{\netfn}{\operatorname{net}}
\providecommand{\etal}{et al.\@}
\section{Experimental Details}
\label{sec:exp}

\subsection{Backbone and Training}
\label{sec:exp:backbone}

\begin{figure}
\centering
\includegraphics[width=\linewidth]{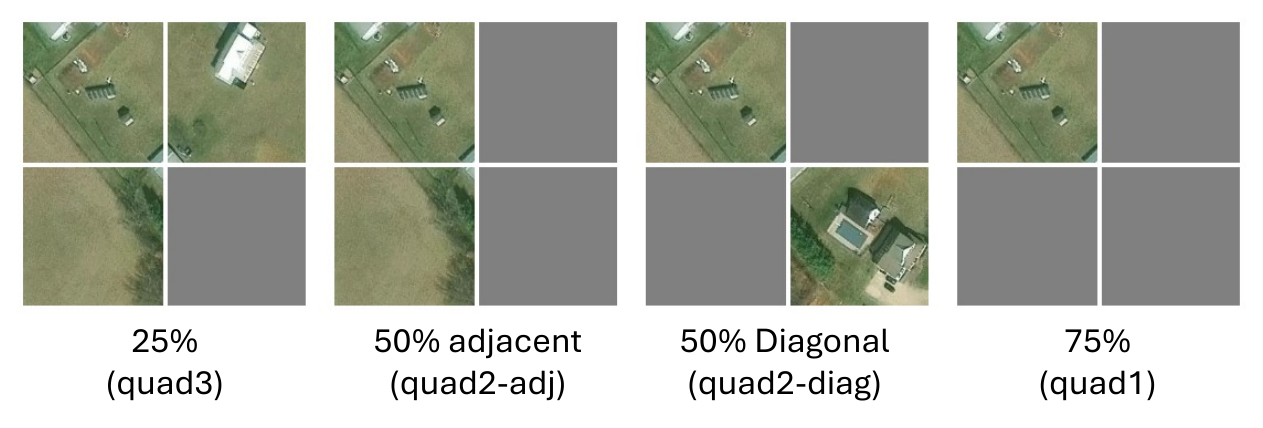}
\caption{Single-tile outpainting training and evaluation regimes.}
\Description{Diagram of the outpainting mask regimes used for training and evaluation, including quad3, quad2 adjacent, quad2 diagonal, and quad1 masks.}
\label{fig:outpainting_eval}
\end{figure}

Both operators use a \textsc{JiT}-H or \textsc{JiT}-B backbone, with SR models using a patch size of 32 and OP models using a patch size of 16. Training follows the flow objective of \cref{sec:method} with flow time drawn from a logit-normal schedule,
$(\mu, \sigma)=(-0.8,0.8)$. Both operators
follow a four-stage, easy-to-hard curriculum with stage transitions at $20/40/60\%$ of
the step budget (rounded to $5$k). SR stages are defined over parent zoom (\emph{coarse}:
parent zoom $\le 15$): (i) coarse-only, (ii) $50/50$ coarse/fine, (iii) uniform over all
zoom levels, (iv) complete quads only. OP stages are defined over the mask
(\cref{fig:outpainting_eval}): (i) \emph{quad3} (three known quadrants, $25\%$ hole),
(ii) \emph{quad2} (two known, adjacent or diagonal, $50\%$), (iii) \emph{quad1} (one
known, $75\%$), (iv) a uniform mixture of all regimes. All training hyperparameters are listed in \cref{tab:hparams} (Appendix). 
For per-operator evaluation, we sample with a $50$-step Euler solver and guidance scale $1.0$; the
multi-scale engine settings are detailed in \cref{sec:exp:engine}.

\subsection{Operator Instantiation}
\label{sec:exp:operators}

\paragraph{Super-resolution.} The semantic stream uses a frozen, satellite-pretrained \textsc{DINOv3} ViT-L \cite{simoni2025dinov3}; conditioning dropout is $p_{\text{drop}}{=}0.1$. The main SR benchmark (\cref{tab:sr-results}) samples from pure noise; a bicubic warm start, which initializes from the parent's bicubic $2\times$ upsample noised to an intermediate $\tau$, is used only for the SR ablation (\cref{tab:sr-ablation}, $\tau=0.5$, total steps= 500k) and for the pyramid engine's SR pass (\cref{sec:exp:engine}, $\tau=0.4$). The auxiliary per-quadrant, $\tau$-weighted LPIPS loss (\cref{eq:srloss}) uses weight $\lambda_{\mathrm{perc}}{=}0.5$.

\paragraph{Outpainting.} Cross-attention is restricted to patches that are more than half known. Conditioning dropout is $0.1$, as in SR. The $7$-channel context and zoom embedding follow \cref{sec:method:op}.

\subsection{Baseline Setup}
\label{sec:appendix-baselines}

For SR, we evaluate against strong super-resolution baselines FastDiffSR~\citep{meng2024fastdiffsr},
SwinIR~\citep{liang2021swinir}, and ZoomLDM~\citep{yellapragada2025zoomldm} under the
same Git-10M parent-to-child scoring protocol used for \genesis. 
Each method produces the target child tile at the requested child zoom, scored against the held-out child tile with the metrics in \cref{tab:sr-results}.

For OP, we compare against SD2-Inpaint and its text-conditioned variant
SD2-Inpaint+Text~\citep{rombach2022high}, and Text2Earth-Inpaint~\citep{liu2025text2earth},
on the same $256{\times}256$ tile masks used for \genesis. 
All OP methods are scored with the same hole-aware metrics and the boundary-difference (B-diff) measure reported in \cref{tab:op-results}.

\subsection{Pyramid Generation}
\label{sec:exp:engine}

In this section, we provide the system-level details that were deferred in
\cref{sec:method:engine}; remaining hyperparameters are in \cref{tab:hparams}.

\paragraph{Seamless super-resolution.} The SR pass over the working level
(\cref{alg:genesis}, step~3) does not super-resolve tiles independently, which would create seams at tile boundaries. Instead, we run a MultiDiffusion-style trajectory over the whole
stitched mosaic, warm-started at $\tau{=}0.4$ from its bicubic upsample: overlapping
$256{\times}256$ windows (overlap $32$\,px) are denoised in parallel, and their per-step
velocities are blended across the canvas with a plateau weighting (close to $1$ over
window interiors, ramping down across overlaps), so the mosaic is upsampled seam-free in
a single trajectory. Leaf seeds are re-anchored at every step at the correct noise level.

\paragraph{Maximum coverage outpainting.} We use offset $r{=}1$, so the working level is the
penultimate level. To complete it (step~2) we treat each
$2{\times}2$ block of $128{\times}128$ quadrants as one $256{\times}256$ OP problem. Each
round selects a maximal set of cell-disjoint, most-constrained windows so they can be
filled in one batched sampler pass, and writes completed tiles back as soon as their four
quadrants are resolved.

\paragraph{Seam repair.} After the SR pass, we score every internal leaf boundary by its mean pixel discontinuity and, worst-first, repair the top boundaries (\cref{alg:genesis}, step~4; threshold $14$, up to $64$ boundaries) with a greedy OP pass: $256{\times}256$ window centered on the seam, with a masked band of width $64$\, px over
the seam. These OP operations repair seams in order of pixel-discontinuity score, so each additional
operation further harmonizes the mosaic. The regenerated band is blended back using a soft-edged (blurred) mask, so it transitions smoothly into the surrounding, unchanged pixels.

\paragraph{Reconciliation.} A bottom-up downsampling sweep (\cref{alg:genesis}, step~5) makes
coarser tiles consistent with the leaves by replacing each parent with the downsampled
mosaic of its children. We apply this only where needed: a working-level tile is
re-derived from its children only if seam repair modified them, and otherwise keeps its
original outpainted content, which is sharper than a downsample of the super-resolved
leaves.

\begin{table*}[th!]
\centering
\scriptsize
\setlength{\tabcolsep}{2pt}
\caption{Super-resolution results on Git-10M random\_test and spatial\_test. We report reconstruction (PSNR, SSIM), perceptual (LPIPS, DISTS), and distributional (FID) metrics for each child zoom group. Higher is better for PSNR and SSIM, lower is better for LPIPS, DISTS, and FID.}
\label{tab:sr-results}
\resizebox{\textwidth}{!}{%
\begin{tabular}{@{}llccccc|ccccc|ccccc@{}}
\toprule
Split   & Model                                           & \multicolumn{5}{c|}{child z12--15}                                                             & \multicolumn{5}{c|}{child z16--17}                                                             & \multicolumn{5}{c}{child z18}                                                                  \\ \cmidrule(l){3-17} 
        &                                                 & PSNR $\uparrow$ & SSIM $\uparrow$ & LPIPS $\downarrow$ & DISTS $\downarrow$ & FID $\downarrow$ & PSNR $\uparrow$ & SSIM $\uparrow$ & LPIPS $\downarrow$ & DISTS $\downarrow$ & FID $\downarrow$ & PSNR $\uparrow$ & SSIM $\uparrow$ & LPIPS $\downarrow$ & DISTS $\downarrow$ & FID $\downarrow$ \\ \midrule
random  & Bicubic                                         & \textbf{27.59}  & \textbf{0.775}  & 0.207              & 0.167              & 31.21            & \textbf{22.56}  & \textbf{0.632}  & 0.304              & 0.201              & 33.51            & \textbf{22.68}  & \textbf{0.630}  & 0.298              & 0.185              & 22.61            \\
random  & FastDiffSR~\citep{meng2024fastdiffsr}           & 23.78           & 0.664           & 0.214              & 0.211              & 40.19            & 20.50           & 0.553           & 0.286              & 0.245              & 43.79            & 20.57           & 0.554           & 0.274              & 0.230              & 36.58            \\
random  & SwinIR~\citep{liang2021swinir}                  & 26.70           & 0.760           & 0.159              & 0.156              & 39.22            & 22.04           & 0.615           & 0.244              & 0.192              & 47.23            & 22.14           & 0.615           & 0.229              & 0.179              & 32.07            \\
random  & ZoomLDM~\citep{yellapragada2025zoomldm} & 22.25           & 0.535           & 0.274              & 0.212              & 70.10            & 19.77           & 0.475           & 0.279              & 0.225              & 50.97            & 19.73              & 0.460              & 0.275                 & 0.224                 & 42.38               \\
random  & \genesis-B                             & 27.37           & 0.767           & 0.150              & 0.131              & 17.16            & 22.52           & 0.618           & 0.233              & 0.169              & 16.81            & 22.57           & 0.625           & 0.222              & 0.160              & 10.30            \\
random  & \genesis-H                             & 27.17           & 0.762           & \textbf{0.129}     & \textbf{0.124}     & \textbf{14.54}   & 22.53           & 0.613           & \textbf{0.199}     & \textbf{0.158}     & \textbf{14.85}   & 22.41           & 0.619           & \textbf{0.188}     & \textbf{0.151}     & \textbf{9.41}    \\ \midrule
spatial & Bicubic                                         & \textbf{27.66}  & \textbf{0.774}  & 0.207              & 0.167              & 29.40            & 22.44           & \textbf{0.629}  & 0.305              & 0.202              & 32.82            & \textbf{23.05}  & \textbf{0.630}  & 0.296              & 0.184              & 22.67            \\
spatial & FastDiffSR~\citep{meng2024fastdiffsr}           & 23.83           & 0.664           & 0.213              & 0.210              & 37.11            & 20.40           & 0.550           & 0.288              & 0.246              & 44.66            & 20.92           & 0.553           & 0.273              & 0.232              & 38.67            \\
spatial & SwinIR~\citep{liang2021swinir}                  & 26.78           & 0.759           & 0.160              & 0.156              & 37.21            & 21.92           & 0.612           & 0.247              & 0.193              & 46.79            & 22.49           & 0.615           & 0.232              & 0.179              & 32.56            \\
spatial & ZoomLDM~\citep{yellapragada2025zoomldm} & 22.31           & 0.534           & 0.276              & 0.213              & 69.11            & 19.69           & 0.472           & 0.280              & 0.226              & 51.16            & 19.99              & 0.460              & 0.277                 & 0.226                 & 43.45               \\
spatial & \genesis-B                             & 27.45           & 0.768           & 0.149              & 0.131              & 15.61            & 22.41           & 0.614           & 0.234              & 0.169              & 16.38            & 22.91           & 0.623           & 0.221              & 0.160              & 10.36            \\
spatial & \genesis-H                             & 27.25           & 0.763           & \textbf{0.128}     & \textbf{0.124}     & \textbf{13.08}   & \textbf{22.44}  & 0.609           & \textbf{0.200}     & \textbf{0.158}     & \textbf{14.75}   & 22.73           & 0.616           & \textbf{0.186}     & \textbf{0.149}     & \textbf{9.23}    \\ \bottomrule
\end{tabular}}
\end{table*}

\begin{table*}[t!]
\centering
\scriptsize
\setlength{\tabcolsep}{2pt}
\caption{Outpainting results on Git-10M random\_test and spatial\_test. We report LPIPS, DISTS, a boundary difference metric (B-diff), and FID for each mask regime. Lower is better for all metrics.}
\label{tab:op-results}
{\renewcommand{\arraystretch}{1.05}
\resizebox{\textwidth}{!}{%
\begin{tabular}{ll*{4}{ccccc}}
\toprule
Split & Model & \multicolumn{4}{c}{quad3} & \multicolumn{4}{c}{quad2\_adj} & \multicolumn{4}{c}{quad2\_diag} & \multicolumn{4}{c}{quad1} \\
\cmidrule(lr){3-6} \cmidrule(lr){7-10} \cmidrule(lr){11-14} \cmidrule(lr){15-18}
& & LPIPS $\downarrow$ & DISTS $\downarrow$ &  B-diff $\downarrow$ & FID $\downarrow$ & LPIPS $\downarrow$ & DISTS $\downarrow$ &  B-diff $\downarrow$ & FID $\downarrow$ & LPIPS $\downarrow$ & DISTS $\downarrow$ &  B-diff $\downarrow$ & FID $\downarrow$ & LPIPS $\downarrow$ & DISTS $\downarrow$ &  B-diff $\downarrow$ & FID $\downarrow$ \\
\midrule
random  & SD2-Inpaint~\citep{rombach2022high}          & 0.398              & 0.307              & 33.01               & 5.65             & 0.449              & 0.337              & 40.37               & 15.11            & 0.398              & 0.306              & 32.75               & 14.16            & 0.469              & 0.351              & 43.93               & 31.90            \\
random  & SD2-Inpaint+Text~\citep{rombach2022high}     & 0.376              & 0.297              & 32.10               & 4.72             & 0.422              & 0.326              & 39.13               & 13.70            & 0.378              & 0.298              & 32.04               & 12.74            & 0.438              & 0.340              & 42.45               & 30.86            \\
random  & Text2Earth-Inpaint~\citep{liu2025text2earth} & 0.332              & 0.276              & 31.91               & 3.77             & 0.367              & \textbf{0.297}     & 36.82               & 9.46             & 0.334              & 0.277              & 31.87               & 8.95             & 0.391              & \textbf{0.310}     & 39.64               & 19.34            \\
random  & \genesis-B                                 & 0.356              & 0.293              & 31.59               & 4.64             & 0.391              & 0.314              & 36.15               & 13.82            & 0.358              & 0.293              & 31.55               & 14.72            & 0.409              & 0.327              & 38.70               & 33.76            \\
random  & \genesis-H                                 & \textbf{0.326}     & \textbf{0.275}     & \textbf{30.53}      & \textbf{3.23}    & \textbf{0.364}     & 0.298              & \textbf{35.20}      & \textbf{8.20}    & \textbf{0.327}     & \textbf{0.275}     & \textbf{30.46}      & \textbf{8.05}    & \textbf{0.385}     & 0.312              & \textbf{38.00}      & \textbf{16.99}   \\ \midrule
spatial & SD2-Inpaint~\citep{rombach2022high}          & 0.396              & 0.308              & 32.64               & 6.01             & 0.447              & 0.339              & 40.28               & 16.42            & 0.395              & 0.307              & 32.36               & 15.13            & 0.468              & 0.353              & 43.77               & 33.91            \\
spatial & SD2-Inpaint+Text~\citep{rombach2022high}     & 0.372              & 0.297              & 31.49               & 4.85             & 0.418              & 0.328              & 38.76               & 14.39            & 0.374              & 0.298              & 31.56               & 13.26            & 0.434              & 0.341              & 41.85               & 31.26            \\
spatial & Text2Earth-Inpaint~\citep{liu2025text2earth} & 0.328              & 0.276              & 31.24               & 3.73             & 0.364              & \textbf{0.297}     & 36.29               & 9.58             & 0.331              & 0.277              & 31.27               & 9.07             & 0.386              & \textbf{0.311}     & 38.97               & 19.91            \\
spatial & \genesis-B                                 & 0.349              & 0.291              & 30.76               & 4.65             & 0.384              & 0.313              & 35.44               & 13.89            & 0.352              & 0.293              & 30.85               & 14.70            & 0.401              & 0.325              & 37.82               & 33.68            \\
spatial & \genesis-H                                 & \textbf{0.321}     & \textbf{0.274}     & \textbf{29.89}      & \textbf{3.24}    & \textbf{0.358}     & \textbf{0.297}     & \textbf{34.52}      & \textbf{8.11}    & \textbf{0.321}     & \textbf{0.275}     & \textbf{29.83}      & \textbf{7.90}    & \textbf{0.379}     & \textbf{0.311}     & \textbf{37.10}      & \textbf{17.23}   \\ 

\bottomrule
\end{tabular}}}
\end{table*}

\subsection{Evaluation}

We evaluate the individual \genesis operators on our Git-10M~\citep{liu2025text2earth} test splits, and the full pyramid-completion engine under different seed conditions on our \densedataset benchmark.

\paragraph{Seed protocols. } We evaluate three seeding regimes that vary the
number and spatial placement of seeds, each applied to all $500$ subtrees ($1{,}500$ pyramid generations total):
\begin{enumerate}[label=(\Alph*)]
\item \textbf{Single seed.} One seed per subtree, its level drawn with weight proportional to its tile count (favoring finer, more realistic seeds). A tile is then chosen uniformly at that level, and results are stratified by the drawn level. This protocol tests full-pyramid generation from a single anchor.
\item \textbf{Three independent seeds.} One seed at each non-root level, none an ancestor or descendant of another. This protocol tests reconciliation of multi-scale anchors at unrelated locations.
\item \textbf{Four spatially independent seeds.} Two leaf, one intermediate, and one coarse seed, none an ancestor of another. This protocol tests dense leaf anchors under coarse guidance.
\end{enumerate}

\paragraph{Super-resolution.} The model generates the child tiles from a parent; results
are grouped by child zoom range ($z12$--$15$, $z16$--$17$, $z18$). The number of tiles
per group is 2470/2003/4096 for the random test split and 3146/2087/4096 for the spatial
test split. We report PSNR, SSIM~\cite{wang2004image}, LPIPS~\cite{zhang2018unreasonable}, DISTS~\cite{ding2020iqa}, and FID~\cite{heusel2017gans} over the full generated tile\footnote{Reported LPIPS uses a SqueezeNet backbone; the training-loss LPIPS of \cref{eq:srloss} uses AlexNet, and FID uses Inception-v3.}.

\paragraph{Outpainting.} The model completes masked regions of a $256{\times}256$ tile
under four regimes (\texttt{quad3}, \texttt{quad2\_adj}, \texttt{quad2\_diag},
\texttt{quad1}; \cref{fig:outpainting_eval}), evaluated on both \texttt{random\_test} and
\texttt{spatial\_test} (\cref{sec:appendix:splits}). We evaluate $8{,}192$ tiles per split and regime. LPIPS and DISTS are \emph{hole-aware}, computed only over the synthesized region (decomposed into quadrants when the hole aligns to them); FID is computed on the full composited tile, following standard inpainting practice. We also introduce a boundary-difference metric (B-diff) to evaluate how seamlessly the generated region blends into the surrounding image. B-diff calculates the absolute pixel-value difference between the generated and real pixels within a 4-pixel-wide ring along the hole boundary.

\paragraph{Pyramid generation.} On \densedataset we report three groups of metrics: image
quality (PSNR, SSIM, LPIPS, FID); semantic alignment between the generated and
ground-truth regions, both embedding-based (CLIP-I~\cite{radford2021learning},
DINOv3-sat~\cite{simoni2025dinov3}) and captioning-based (Caption); and pyramid
consistency, the within- and cross-level reconstruction metrics LR-PSNR$_\text{box}$ and
xLR-PSNR together with the spectral metric RAPSD~\cite{lugmayr2022ntire, lai2017deep, ulichney1987digital}. Unlike standard single-image metrics, which score each tile in isolation, the pyramid-consistency metrics score relations between zoom levels: parent-child self-consistency (LR-PSNRbox), cross-scale fidelity to ground truth (xLR-PSNR), and spectral realism (RAPSD). Our goal is a plausible, seed-consistent pyramid rather than pixel-exact reconstruction, so the distributional, semantic, and consistency metrics are primary; reconstruction metrics are reported for completeness. Further details on the semantic alignment and
pyramid-consistency metrics are in Appendix Sections \ref{app:semantic} and \ref{app:pyramid-metrics}.

\begin{table*}[t!]
\centering
\scriptsize
\setlength{\tabcolsep}{2pt}
\caption{Evaluation on \densedataset under seed protocols A-C (A: 1 seed, hardest; C: 4 seeds, easiest) and their aggregate (All) for the task of multi-scale tile completion.}
\label{tab:dense500-results}
\resizebox{\textwidth}{!}{%
\begin{tabular}{llcccccccccc}
\toprule
& & \multicolumn{4}{c}{Image quality} & \multicolumn{3}{c}{Semantic alignment} & \multicolumn{3}{c}{Pyramid consistency} \\
 \cmidrule(lr){3-6} \cmidrule(lr){7-9} \cmidrule(lr){10-12}
Model & Protocol & PSNR\,$\uparrow$ & SSIM\,$\uparrow$ & LPIPS\,$\downarrow$ & FID\,$\downarrow$ & CLIP-I\,$\uparrow$ & DINOv3-sat\,$\uparrow$ & Caption\,$\uparrow$ & LR-PSNR$_{\mathrm{box}}$\,$\uparrow$ & xLR-PSNR\,$\uparrow$ & RAPSD\,$\downarrow$ \\
\midrule
\genesis-B & All         & 15.99            & 0.271            & 0.455               & 120.1             & 0.800              & 0.363                  & 0.783               & 35.11                                & 16.54                & 1.121               \\
         & A (1 seed)  & 15.65            & 0.262            & 0.497               & 131.7             & 0.789              & 0.336                  & 0.777               & 35.93                                & 15.93                & 1.795               \\
         & B (3 seeds) & 16.13            & 0.274            & 0.436               & 120.8             & 0.805              & 0.374                  & 0.782               & 34.70                                & 16.79                & 0.808               \\
         & C (4 seeds) & 16.21            & 0.277            & 0.431               & 122.4             & 0.807              & 0.380                  & 0.791               & 34.70                                & 16.90                & 0.760               \\ \midrule
\genesis-H & All         & 16.95            & 0.298            & 0.423               & 72.0              & 0.843              & 0.419                  & 0.820               & 37.19                                & 17.59                & 1.281               \\
         & A (1 seed)  & 15.96            & 0.277            & 0.495               & 99.4              & 0.808              & 0.380                  & 0.793               & 37.53                                & 16.25                & 2.001               \\
         & B (3 seeds) & 17.43            & 0.308            & 0.390               & 64.4              & 0.858              & 0.434                  & 0.822               & 37.05                                & 18.21                & 0.965               \\
         & C (4 seeds) & 17.49            & 0.309            & 0.383               & 62.2              & 0.864              & 0.444                  & 0.846               & 36.98                                & 18.30                & 0.878               \\ 
\bottomrule
\end{tabular}
}
\end{table*}

\begin{figure}[t!]
  \includegraphics[width=0.43\textwidth]{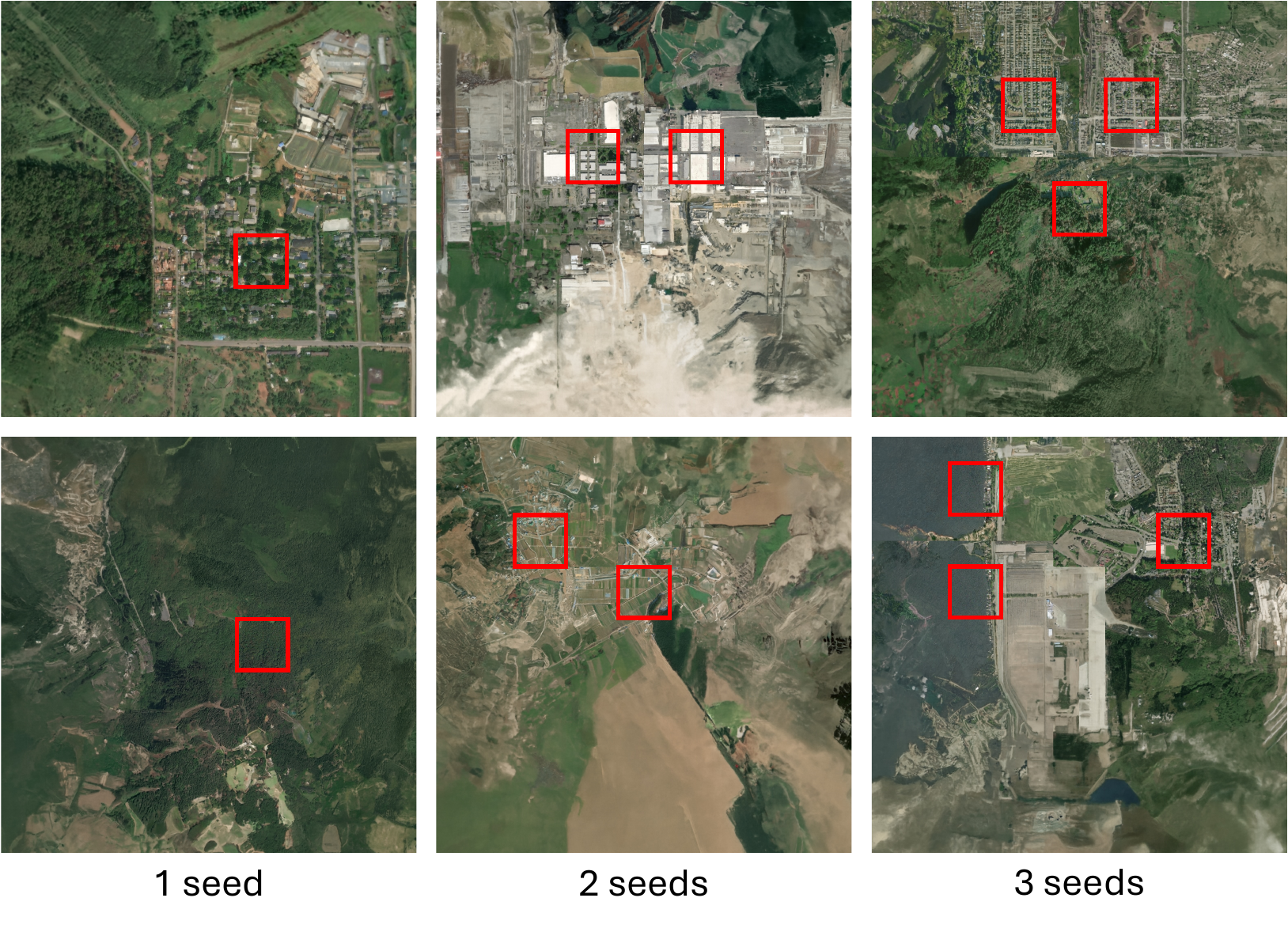}
\caption{\genesis outpainting results at zoom level 16 across three sites, each initialized from 1–3 seed tiles (\textcolor{red}{red}).}
  \label{fig:outpainting_examples}
\end{figure}

\section{Results}
\label{sec:results}

\subsection{Super-resolution}
\label{sec:evaluation-sr}

\cref{tab:sr-results} reports super-resolution results on both splits, \texttt{random\_test}
and \texttt{spatial\_test}, each stratified into three child-zoom groups.
\genesis-H is best on every perceptual and distributional metric, holding the lowest LPIPS, DISTS, and FID in all six
split/zoom groups. Its FID stays in the $9.2$--$14.8$ range, against $32$--$47$ for SwinIR
and $42$--$70$ for ZoomLDM, and its LPIPS is roughly
$20\%$ lower than SwinIR's in every group (e.g.\ $0.129$ vs.\ $0.159$ at $z12$--$15$).
Performance is nearly identical across the two test splits, indicating that quality does not depend on whether the held-out tiles are urban-
balanced or uniformly sampled. Difficulty also varies with scale: reconstruction metrics
(PSNR, SSIM) are highest for the coarsest children ($z12$--$15$), while FID is consistently
lowest at $z18$, where fine texture dominates, and the perceptual gap to baselines is widest.
Bicubic scores highest on PSNR and SSIM almost everywhere, but only because interpolation
preserves low-frequency structure; its blurry output is penalized sharply by LPIPS and FID,
showing that the pixel-fidelity metrics PSNR and SSIM do not reflect perceptual quality. We emphasize that the SR task represents a fundamental operation in hierarchical pyramid generation, which must be performed many times in order to traverse the resolution hierarchy. As such, errors from a weaker operator propagate, and small differences in fidelity and coherence become magnified across scales. \genesis significantly advances super-resolution performance over the second-best baseline, enabling generation across arbitrary hierarchical spans.

\begin{table}[h!]
\centering
\scriptsize
\setlength{\tabcolsep}{2pt}
\caption{Super-resolution ablation on \texttt{val\_spatial} with 5000 samples under the bicubic-init SR setting. PSNR and SSIM are higher-is-better; LPIPS, DISTS, and FID are lower-is-better.}
\label{tab:sr-ablation}
\resizebox{0.475\textwidth}{!}{%
\begin{tabular}{@{}ccrrrrr@{}}
\toprule
DINOv3   & $\mathcal{L}_{LPIPS}$  & PSNR $\uparrow$  & SSIM $\uparrow$ & LPIPS $\downarrow$ & DISTS $\downarrow$ & FID $\downarrow$ \\ \midrule
\xmark & \xmark & 24.18          & 0.669          & 0.231             & 0.167             & 13.46         \\
\cmark & \xmark & \textbf{24.23} & \textbf{0.673} & 0.223             & 0.160             & 12.74          \\
\xmark & \cmark & 24.14          & 0.663          & 0.212             & 0.158             & 11.12          \\
\cmark & \cmark & 24.15          & 0.664          & \textbf{0.204}    & \textbf{0.152}    & \textbf{10.30} \\
\bottomrule
\end{tabular}}
\end{table}

\cref{tab:sr-ablation} ablates the two optional SR components on \texttt{val\_spatial}.
Each helps, in complementary ways. DINOv3 conditioning gives small gains across the board
and the best PSNR/SSIM, consistent with a semantic prior that stabilizes structure. The
LPIPS loss drives the larger perceptual improvement, further reducing FID at a slight cost to PSNR/SSIM. Combining the two yields the best LPIPS, DISTS, and
FID while keeping PSNR/SSIM essentially unchanged. Guidance in both generated semantics and perceptual quality enables the strong performance needed for hierarchical consistency and distributional correctness in the pyramid generation task.

\subsection{Outpainting}
\label{sec:evaluation-op}

\cref{tab:op-results} reports outpainting across both splits and the four mask regimes.
\genesis-H attains the lowest FID, LPIPS, and boundary difference (B-diff) in all eight
split/regime cells, outperforming SD2-Inpaint, SD2-Inpaint+Text, and Text2Earth-Inpaint on
each. Difficulty scales with hole size: FID rises from about $3.2$ at \texttt{quad3}
($25\%$ hole) to $8$ at \texttt{quad2} ($50\%$) and $17$ at \texttt{quad1} ($75\%$), as less
context remains to anchor the completion. The B-diff advantage matters most for our setting,
since low seam error at the hole boundary is what lets \genesis chain completions into
seam-free extents (\cref{fig:pyramid_example}). On DISTS, \genesis-H is on par
with the strongest baseline, Text2Earth-Inpaint, matching or slightly beating it in most
regimes. As in super-resolution, results are nearly identical across the \texttt{random\_test}
and \texttt{spatial\_test} splits. Like super-resolution, outpainting is a fundamental operation needed for lateral pyramid filling. As outpainting must be chained multiple times, it is essential that the underlying model produces distributionally correct, spatially coherent imagery, free of artifacts and seams. Across all metrics, \genesis is consistently state-of-the-art, providing the properties needed for composing multiple outpainting operations. A set of qualitative examples of the outpainting capability of \genesis is shown in \cref{fig:outpainting_examples}.

\subsection{Summary across Regimes}

\cref{tab:average_fid_summary} aggregates mean FID over all cells (six for SR,
eight for OP). On SR, \genesis-H and \genesis-B achieve the two best mean FIDs ($12.6$ and
$14.4$), significantly ahead of the next-best baseline, bicubic ($28.7$), and far ahead
of model-based SR approaches SwinIR ($39.2$), FastDiffSR ($40.2$), and ZoomLDM ($54.5$). A notable result from this analysis is that no model-based baseline is able to outperform the simple, training-free bicubic upsampling baseline. Intuitively, this indicates that current models are poorly suited to multi-resolution hierarchical synthesis, for which SR more powerful than bicubic upsampling is needed to traverse wide zoom-level ranges.

On OP, \genesis-H is the best
($9.1$), ahead of Text2Earth-Inpaint ($10.5$), with the remaining methods clustered higher
($15.7$--$17.3$). A single model family thus leads both subtasks, which is what makes
composing the two operators into one engine worthwhile.

\begin{figure*}[t!]
  \includegraphics[width=0.9\textwidth]{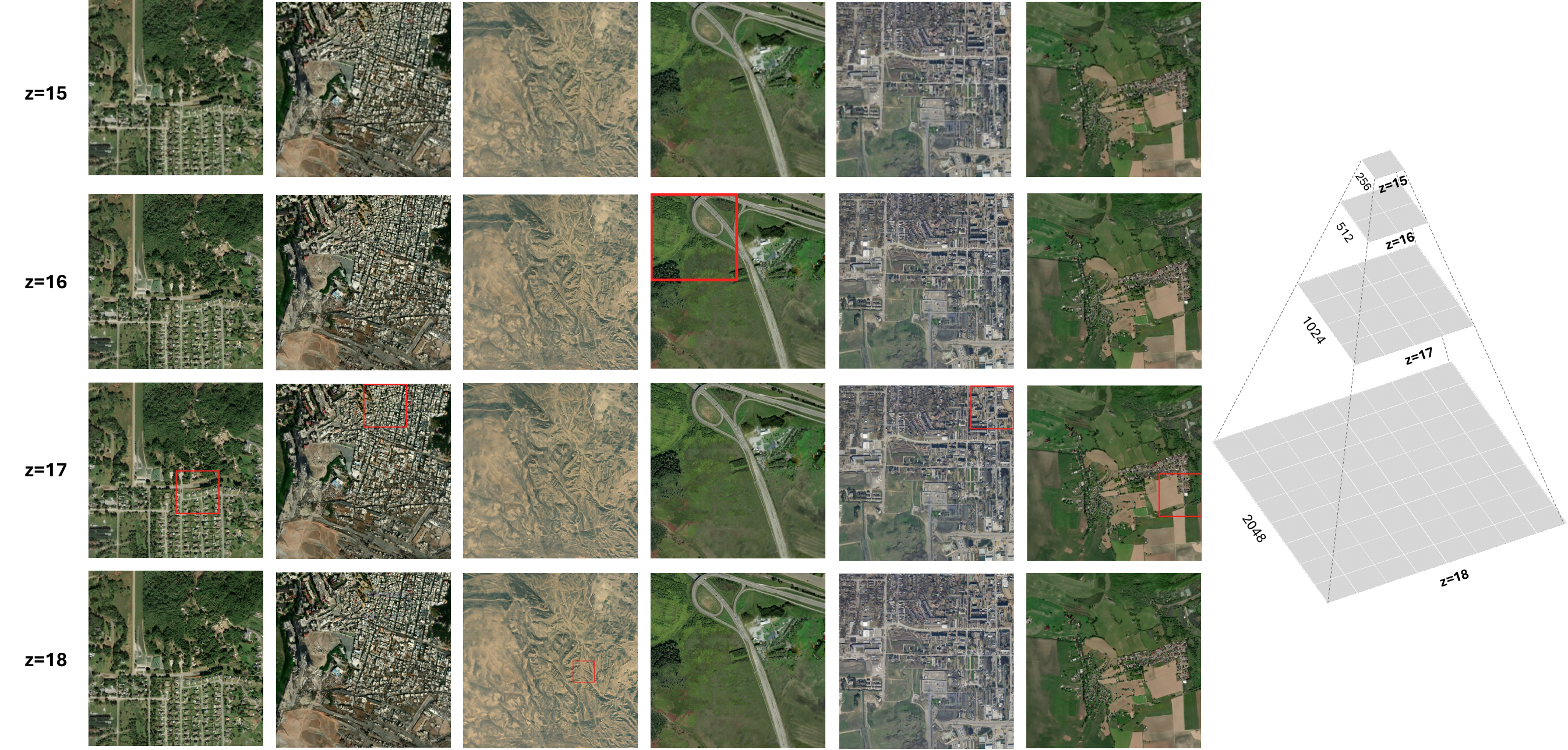}
\caption{\genesis pyramid completions across many sites from seed tiles
(\textcolor{red}{red}). Each column is a distinct site, and each row a zoom level
($z{=}15$ to $z{=}18$); the schematic (right) shows the underlying quadtree. Outputs are
consistent vertically across zoom levels and horizontally within each level.}
  \label{fig:qual_figure_collage}
\end{figure*}

\begin{table}[h!]
\centering
\small
\caption{Mean FID across all evaluation regimes.}
\label{tab:average_fid_summary}
\begin{tabular}{llr}
\toprule
Task & Method & Avg. FID $\downarrow$ \\
\midrule
SR & \genesis-H & \textbf{12.642} \\
SR & \genesis-B & 14.434 \\
SR & Bicubic & 28.703 \\
SR & SwinIR~\citep{liang2021swinir} & 39.180 \\
SR & FastDiffSR~\citep{meng2024fastdiffsr} & 40.167 \\
SR & ZoomLDM~\citep{yellapragada2025zoomldm} & 54.528 \\
\midrule
OP & \genesis-H & \textbf{9.119} \\
OP & Text2Earth-Inpaint~\citep{liu2025text2earth} & 10.477 \\
OP & SD2-Inpaint+Text~\citep{rombach2022high} & 15.722 \\
OP & \genesis-B & 16.734 \\
OP & SD2-Inpaint~\citep{rombach2022high} & 17.288 \\
\bottomrule
\end{tabular}
\end{table}

\subsection{Full Pyramid Generation}
Multi-scale tile completion itself is a new task with no existing end-to-end
method or established pipeline of third-party operators to compare against;
comparisons with existing methods are therefore conducted on the per-operator
subtasks, where \genesis outperforms the compared baselines under identical
protocols (\cref{tab:sr-results,tab:op-results}), and
\cref{tab:average_fid_summary} benchmarks the two \genesis model capacities (B and H) as reference
points. To evaluate full pyramid generation, we report image-quality, semantic-alignment, and pyramid-consistency
metrics on \densedataset (Table~\ref{tab:dense500-results}), which spans five zoom windows from roughly $76$ to
$0.6$\,m/px (\cref{tab:dense500}). \genesis-H improves over \genesis-B on nearly every metric (FID
$72.0$ vs.\ $120.1$, CLIP-I $0.843$ vs.\ $0.800$, DINOv3-sat $0.419$ vs.\ $0.363$), trading compute for
generative quality.

Across the three seed protocols, denser seeding makes completion easier, as expected. For
\genesis-H, FID falls from $99.4$ with a single seed (protocol A) to $64.4$ and $62.2$ with
three and four seeds (protocols B and C), and semantic alignment rises (CLIP-I
$0.808\!\to\!0.864$, Caption $0.793\!\to\!0.846$). The two pyramid-consistency metrics tell
complementary stories. LR-PSNR$_\text{box}$, which measures whether a generated parent agrees
with its own children, stays high and roughly constant across all protocols, showing that the engine produces internally consistent pyramids regardless of
how many seeds it starts from. xLR-PSNR, which compares against ground truth across scales,
instead rises with seeding ($16.2\!\to\!18.3$), indicating that \genesis can effectively utilize increased context to generate more semantically accurate scenes. RAPSD tells the same story from the frequency domain: the spectral distance shrinks steadily as seeding densifies ($2.00$ with one seed to $0.88$ with four for Genesis-H), showing that additional anchors also pull the generated frequency content toward that of real imagery. \cref{fig:pyramid_example}
shows this cross-scale behavior on a single site: zooming from $z{=}15$ to $z{=}18$. 

\subsection{Qualitative Results}
\cref{fig:qual_figure_collage} shows multi-resolution completions across many sites and varying seed
configurations. \genesis produces spatially and hierarchically consistent imagery, with
minimal seams between neighbors, coherent structure across zoom levels, and content that
matches the semantics of the given seeds. Failure cases are largely observed at the single-seed
extreme (protocol A), where large unseen regions must be hallucinated from a single anchor, and the lack of context causes generations to drift away from the ground-truth semantics. \cref{fig:pyramid_example}, \cref{fig:qual_figure_collage}, \cref{fig:4-level-collage}, and \cref{fig:6-level} demonstrate that \genesis cleanly generates semantically and hierarchically consistent imagery across a large span of zoom levels with mixed-resolution seeding.

\section{Conclusion}
\label{sec:conclusion}
We introduced \emph{multi-scale tile completion}, a new generative task that requires
synthesizing complete satellite-image pyramids from sparse observations while remaining
consistent across both geographic space and zoom levels. To address this task, we proposed
\genesis, a generative engine built on two operators we trained for satellite imagery: a
vertical super-resolution model and a horizontal mask-based outpainting model. Each achieves
state-of-the-art results on its respective subtask, and our algorithm composes them to
complete full pyramids from any seed configuration. To evaluate \genesis on this task, we
introduced \densedataset, a fully observed multi-scale pyramid benchmark spanning diverse
geographic regions, together with a suite of pyramid-level metrics, and used them to show
that \genesis produces pyramids that are realistic, semantically aligned with the seeds, and
consistent across scales. Beyond benchmark performance, Genesis turns sparse, opportunistic acquisitions into complete, navigable pyramids, supporting applications such as populating
virtual environments, prototyping urban layouts, and generating multi-resolution training
data for remote sensing models. We hope that our work will serve as a foundation for future research on multi-scale generative modeling of the Earth.

\section{Acknowledgements}
This research used the TGI RAILs advanced compute resource, which is supported by
the National Science Foundation (award OAC-2232860) and Taylor Geospatial.

\bibliographystyle{ACM-Reference-Format}
\bibliography{citations}
\appendix
\section{Appendix}
\label{sec:appendix}

\subsection{Data Splits}
\label{sec:appendix:splits}

Within the georeferenced Git-10M subset, we identify \emph{complete quads} (parents whose four $z{+}1$ children are all present), which form the population for our evaluation splits. We draw four disjoint evaluation sets of complete quads (\cref{tab:splits}): \emph{spatial} test/val, balanced $50/50$ urban/non-urban (urban: tile center within a World Urban Areas polygon~\cite{world_urban_areas}), and \emph{random} test/val, sampled uniformly. The remaining tiles form the training set, which additionally includes \emph{partial} quads (parents with $1$--$3$ children) used by the masked losses.

\begin{figure}[ht]
\centering
\includegraphics[width=\linewidth]{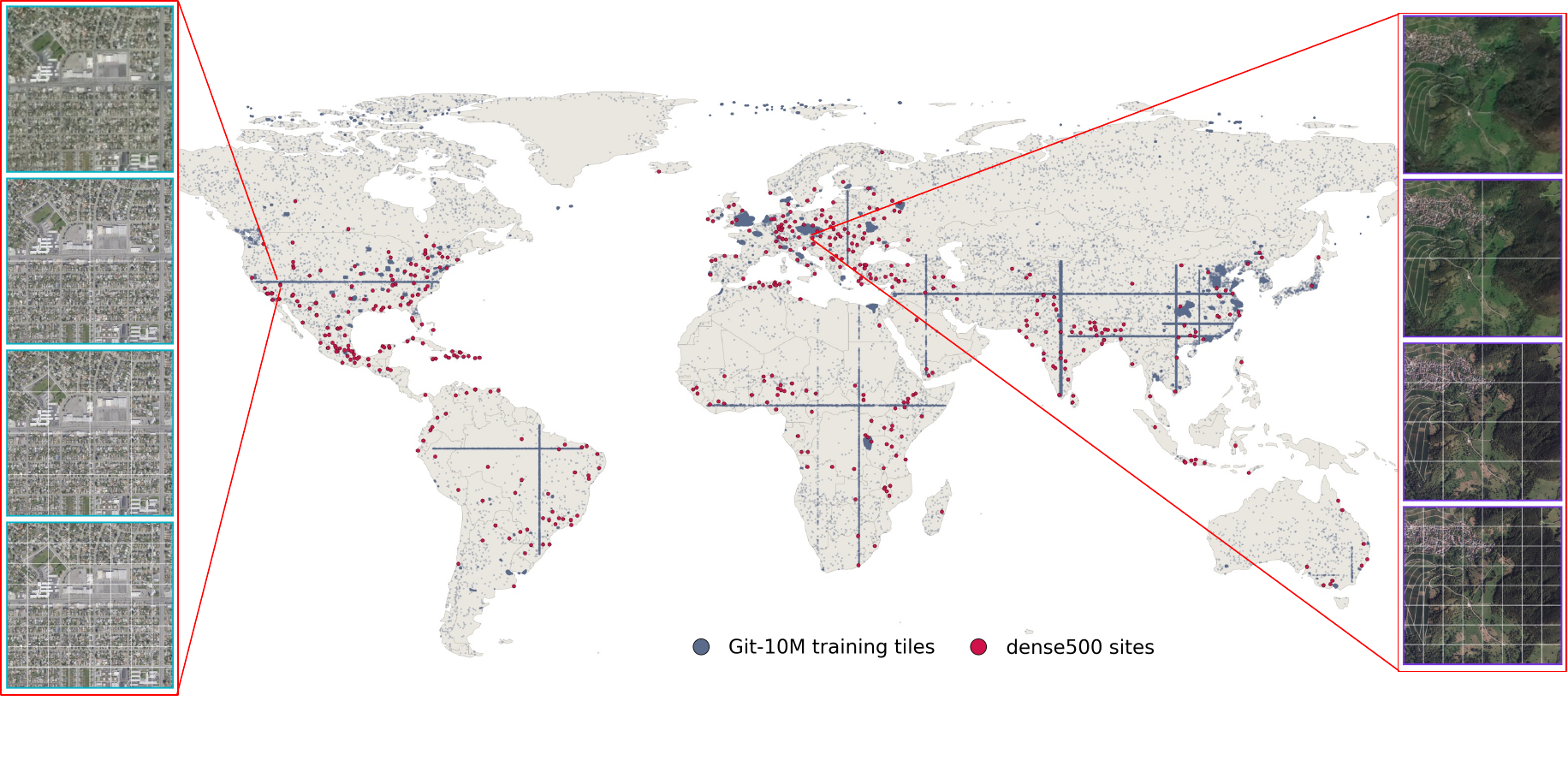}
\caption{Geographic coverage. Git-10M training tiles (small dark markers) sample near-globally; sites are shown in red.}
\Description{Git-10M training tiles (small dark markers) are sampled near-globally; \densedataset sites
    are shown in red.}
\label{fig:data-coverage}
\end{figure}

\begin{table}[h]
\centering
\small
\setlength{\tabcolsep}{6pt}
\caption{Data splits over the georeferenced Git-10M subset.}
\label{tab:splits}
\begin{tabular}{lrr}
\toprule
Split & Complete quads & Tiles \\
\midrule
Train          & $362{,}572$ & $6{,}568{,}890$ \\
Test (spatial) & $10{,}000$  & $49{,}825$      \\
Test (random)  & $10{,}000$  & $49{,}826$      \\
Val (spatial)  & $5{,}000$   & $24{,}958$      \\
Val (random)   & $5{,}000$   & $24{,}969$      \\
\bottomrule
\end{tabular}
\end{table}

\subsection{\densedataset~Site Selection and Filtering}
\label{sec:appendix:dense500}
Site centers are sampled from the LandScan~2024~\cite{lebakula2025landscan} population raster, with an urban restriction defined as cells above the $70$th percentile of populated cells. Each tile is rejected if it is \emph{overzoomed} (upsampled from a coarser native resolution, detected by a low Laplacian variance) or \emph{flat} (e.g.\ open ocean, detected by a low per-channel standard deviation). A site is kept only if all $85$ of its tiles pass, so the published sites contain no upsampled or empty imagery. We observed that the imagery
\emph{source} can change within a single pyramid (satellite vs. aerial layers at fine zooms). This may be due to temporal differences in source capture. The transition produces a visible discontinuity in color and texture between adjacent levels. To keep this source artifact from confounding a benchmark meant to measure generative fidelity across scale, we aggregate an $8{\times}8{\times}8$ RGB color histogram over all tiles at each zoom level and compare adjacent levels with the Bhattacharyya distance and the Pearson correlation of their histograms. A site is rejected if \emph{any} adjacent pair has a Bhattacharyya distance above $0.25$ \emph{and} a Pearson correlation below $0.875$. 
Sites are resampled until $100$ survive for each zoom-window condition as listed in \cref{tab:dense500}.

\subsection{Metric Details}
\label{app:metrics}
We provide additional details on our metrics used for the evaluation of generated pyramids.

\subsection{Semantic Alignment}
\label{app:semantic}

We describe embedding- and captioning-based semantic alignment metrics built on foundation models.

\paragraph{CLIP-I (category level).}
Following the image-image protocol of DreamBooth~\cite{ruiz2023dreambooth}, we
embed the generated and target tiles with a CLIP image encoder and report embedding cosine similarity. 

\paragraph{DINOv3-sat (instance level).}
We also provide embedding similarity metrics from a geospatial foundation model. DINOv3-sat is self-supervised on $493$M satellite
images \cite{simoni2025dinov3}, and provides strong features for Earth observation data. We report embedding cosine similarity on these features as well. 

\paragraph{Caption agreement (scene level).}
We caption both the synthesized region and the corresponding real region with a vision-language model (Qwen2.5-VL-3B-Instruct), embed the two captions with a sentence encoder (Sentence-BERT, all-MiniLM-L6-v2~\cite{reimers2019sentence}), and report the cosine similarity between their representations:
\begin{equation}
  \mathrm{Cap}(\hat{x}, x) \;=\; \cos\!\big(\psi(c(\hat{x})),\, \psi(c(x))\big),
\end{equation} where $c(.)$ denotes the caption and $\psi(.)$ the sentence embedding.

\subsection{Pyramid Consistency}
\label{app:pyramid-metrics}
A correct pyramid must be \emph{internally consistent}, producing a cleanly zoomable multi-resolution structure. This implies parent-child consistency, which we measure by the following metrics:

\paragraph{LR-PSNR (intra-scale self-consistency).}
We measure how well a generated parent agrees with its own generated children,
requiring no ground truth:
\begin{equation}
  \text{LR-PSNR}(p) \;=\; \mathrm{PSNR}\!\big(\,\hat{p},\;\downarrow_2 M(\hat{q})\,\big),
\end{equation} where $\hat{p}$ is the
    generated parent, $\hat{q}$ its four generated children (q the ground-truth children), $M$
    stitches four child tiles into their mosaic, and $\downarrow_2$ denotes 2x downsampling.
We report it under box (area-average) downsampling, the physically correct quadtree operator. Low LR-PSNR means a parent and its children disagree.

\paragraph{xLR-PSNR (cross-scale fidelity).}
Self-consistency can be high while both parent and children are jointly wrong. To
separate consistency from correctness, we compare the generated child mosaic to the
\emph{ground-truth} child mosaic at the parent's resolution,
\begin{equation}
  \text{xLR-PSNR}(p) \;=\;
  \mathrm{PSNR}\!\big(\,\downarrow_2 M(\hat{q}),\;\downarrow_2 M(q)\,\big),
\end{equation}
following the cross-scale LR-PSNR convention~\cite{lai2017deep}. 

\paragraph{RAPSD spectral distance.}
We compute the radially-averaged power spectral density (RAPSD) of the generated and
ground-truth child mosaics and report the mean absolute log-power difference~\cite{durall2020watch},
\begin{equation}
  \text{RAPSD}(p) \;=\;
  \frac{1}{|F|}\sum_{f\in F}\big|\log S_{\hat{q}}(f) - \log S_{q}(f)\big|,
\end{equation}
where $S(f)$ is azimuthally-averaged power at radial frequency $f$.

\begin{table}[t]
\centering
\footnotesize
\setlength{\tabcolsep}{3.5pt}
\begin{tabular}{@{}lcc@{}}
\toprule
 & SR & OP \\
\midrule
\multicolumn{3}{@{}l}{\emph{Architecture}}\\
Backbone                & \textsc{JiT}-B/H  & \textsc{JiT}-B/H \\
Patch size              & $32$              & $16$ \\
Input\,$\to$\,output    & $256\!\to\!512$   & $256\!\to\!256$ \\
\midrule
\multicolumn{3}{@{}l}{\emph{Training}}\\
Optimizer               & \multicolumn{2}{c}{AdamW} \\
$(\beta_1,\beta_2)$     & $(0.9,0.999)$     & $(0.9,0.95)$ \\
Learning rate           & $5\!\times\!10^{-5}$ & $1\!\times\!10^{-5}$ \\
Batch size              & \multicolumn{2}{c}{$256$} \\
Grad.\ clip             & $1.0$             & $0.5$ \\
Precision               & \multicolumn{2}{c}{bf16-mixed} \\
Total steps             & \multicolumn{2}{c}{$800$k} \\
EMA decays              & \multicolumn{2}{c}{$0.9999,\,0.9996$ (eval on $0.9999$ weights)} \\
Noise $(P_{\mu},P_{\sigma})$ & \multicolumn{2}{c}{$(-0.8,\,0.8)$} \\
Curric.\ boundaries     & \multicolumn{2}{c}{$20/40/60\%$ of steps} \\
Cond.\ dropout $p_{\mathrm{drop}}$ & \multicolumn{2}{c}{$0.1$} \\
LPIPS loss weight $\lambda_{\mathrm{perc}}$ & $0.5$ & -- \\
\midrule
\multicolumn{3}{@{}l}{\emph{Inference: per-operator evaluation}}\\
Solver / steps          & \multicolumn{2}{c}{Euler, $50$} \\
CFG scale               & \multicolumn{2}{c}{$1.0$} \\
\midrule
\multicolumn{3}{@{}l}{\emph{Inference: multi-scale tile completion}}\\
Solver / steps          & \multicolumn{2}{c}{Euler, $50$} \\
CFG scale               & $2.5$             & $1.0$ \\
Window / overlap        & $256$, $32$\,px   & -- \\
Warm-start $\tau$       & $0.40$            & -- \\
\bottomrule
\end{tabular}
\caption{Hyperparameters for the two operators, super-resolution (SR) and
outpainting (OP): architecture, training, and the sampler settings used in
each evaluation.}
\label{tab:hparams}
\end{table}

\begin{figure*}
    \centering
    \includegraphics[width=0.95\linewidth]{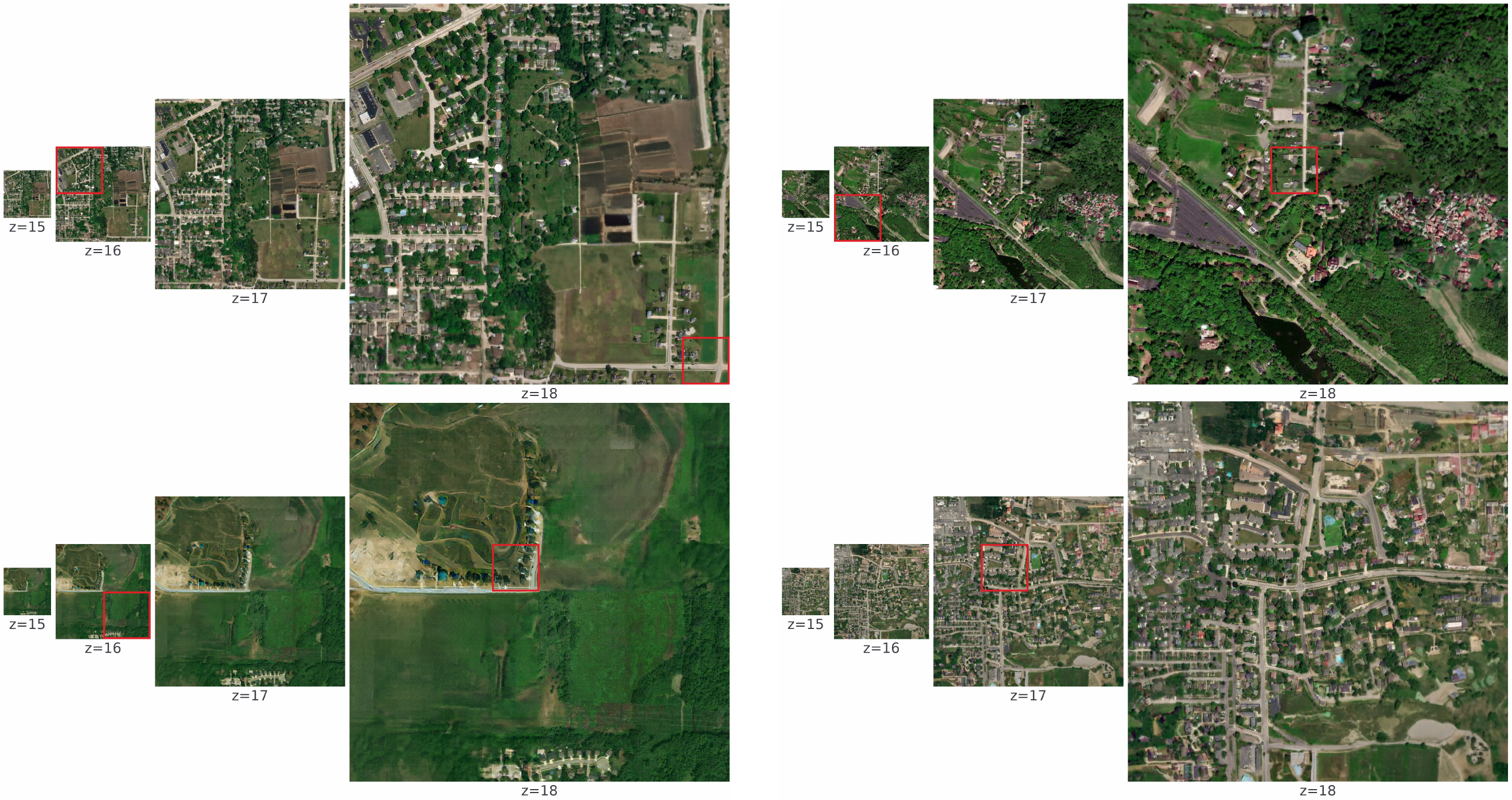}
    \caption{Examples of 4-level pyramids generated using \genesis, conditioned on seed tiles (\textcolor{red}{red}) at different zoom levels.}
    \label{fig:4-level-collage}
\end{figure*}

\begin{figure}
    \centering
    \includegraphics[height=0.475\textheight]{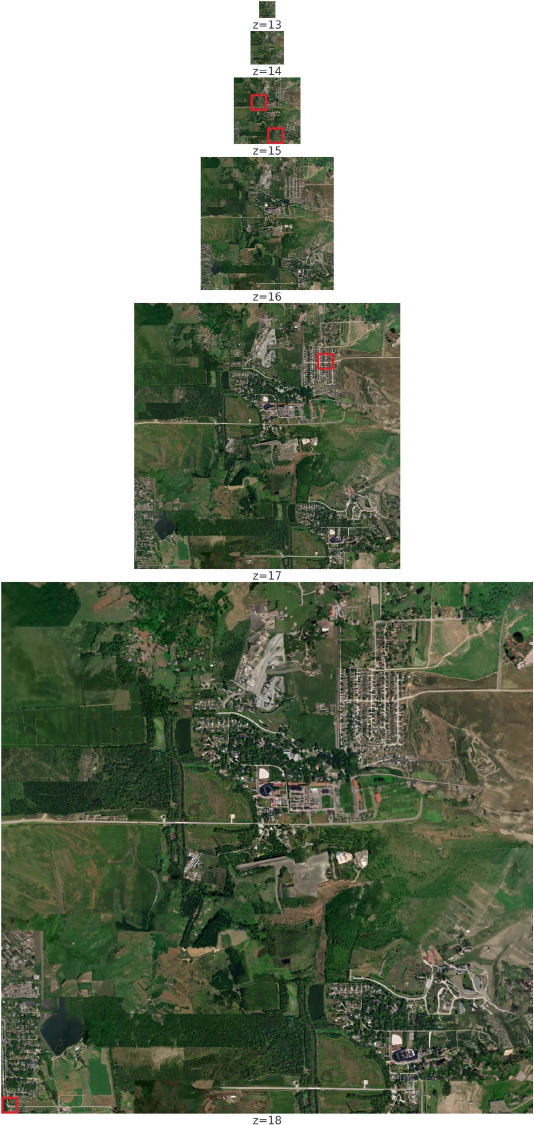}
    \caption{Example of a 6-level pyramid generated using \genesis, conditioned on seed tiles (\textcolor{red}{red}) at different zoom levels.}
    \label{fig:6-level}
\end{figure}

\end{document}